\documentclass[letterpaper, 10 pt, conference]{ieeeconf}  

\IEEEoverridecommandlockouts                              

\usepackage{graphics} 
\usepackage{caption}
\usepackage{subfigure}
\usepackage{epsfig} 
\usepackage{mathptmx} 
\usepackage{times} 
\usepackage{amsmath} 
\usepackage{amssymb}  
\usepackage{array}
\usepackage{url} 
\usepackage{color}
\usepackage{cite}  

\title{\LARGE \bf
	LiDAR Iris for Loop-Closure Detection
}

\author{
	Ying Wang$^{\dag}$,
	Zezhou Sun$^{\dag}$,
	Cheng-Zhong Xu$^{\ddag}$,
	Sanjay E. Sarma$^{\S}$,
	Jian Yang$^{\dag}$, and
	Hui Kong$^{\dag}$
	\thanks{$^{\dag}$ School of Computer Science and Engineering, Nanjing University of Science and Technology, Nanjing, Jiangsu, China.}%
	\thanks{$\ddag$ Department of Computer Science, University of Macau, Macau.}
	\thanks{$^{\S}$ Department of Mechanical Engineering, MIT, Cambridge, MA.}
}

\begin{document}
		
	\maketitle
	\thispagestyle{empty}
	\pagestyle{empty}
	
	\begin{abstract} 
     In this paper,  a global descriptor for a LiDAR point cloud, called LiDAR Iris, is proposed for fast and accurate loop-closure detection. A binary signature image can be obtained for each point cloud after several LoG-Gabor filtering and thresholding operations on the LiDAR-Iris image representation. Given two point clouds, their similarities can be calculated as the Hamming distance of two corresponding binary signature images extracted from the two point clouds, respectively. Our LiDAR-Iris method can achieve a pose-invariant loop-closure detection at a descriptor level with the Fourier transform of the LiDAR-Iris representation if assuming a 3D (x,y,yaw) pose space, although our method can generally be applied to a 6D pose space by re-aligning point clouds with an additional IMU sensor. Experimental results on five road-scene sequences demonstrate its excellent performance in loop-closure detection. 
	\end{abstract}
	
	\section{INTRODUCTION}
		
	During the past years, many techniques have been introduced that can perform simultaneous localization and mapping using both regular cameras and depth sensing technologies based on either structured light,
	ToF, or LiDAR. Even though these sensing technologies are producing accurate depth measurements, however, they are still far from
	perfect. 
	As the existing methods cannot completely eliminate the accumulative error of pose estimation, these methods still suffer from the drift problem. 
	These drift errors could be corrected by incorporating sensing information taken from places that have been visited before, or loop-closure detection, which requires algorithms that are able to recognize revisited areas.  Unfortunately, existing solutions for loop detection for 3D LiDAR data are not both robust and 
	fast enough to meet the demand of real-world SLAM applications. 

\begin{figure}
		\centering 
		\includegraphics[width=0.8\linewidth]{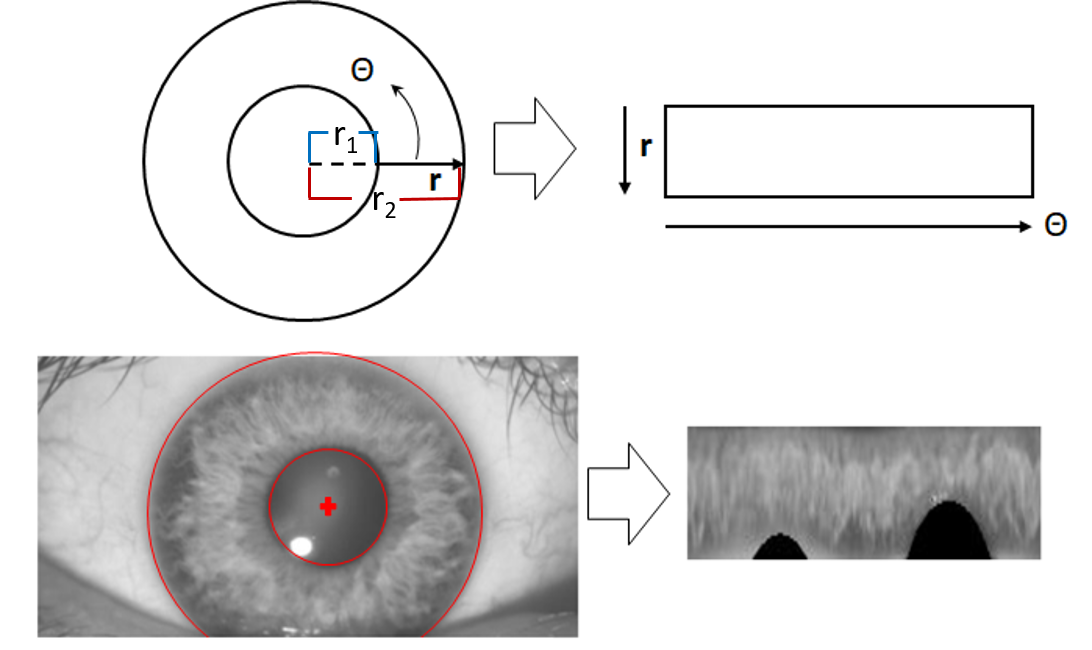}
		\caption{The Daugman's Rubber Sheet Model in person identification based on iris image matching.}
		\label{iris}
	\end{figure}

\begin{figure}
		\centering 
		\includegraphics[width=1\linewidth]{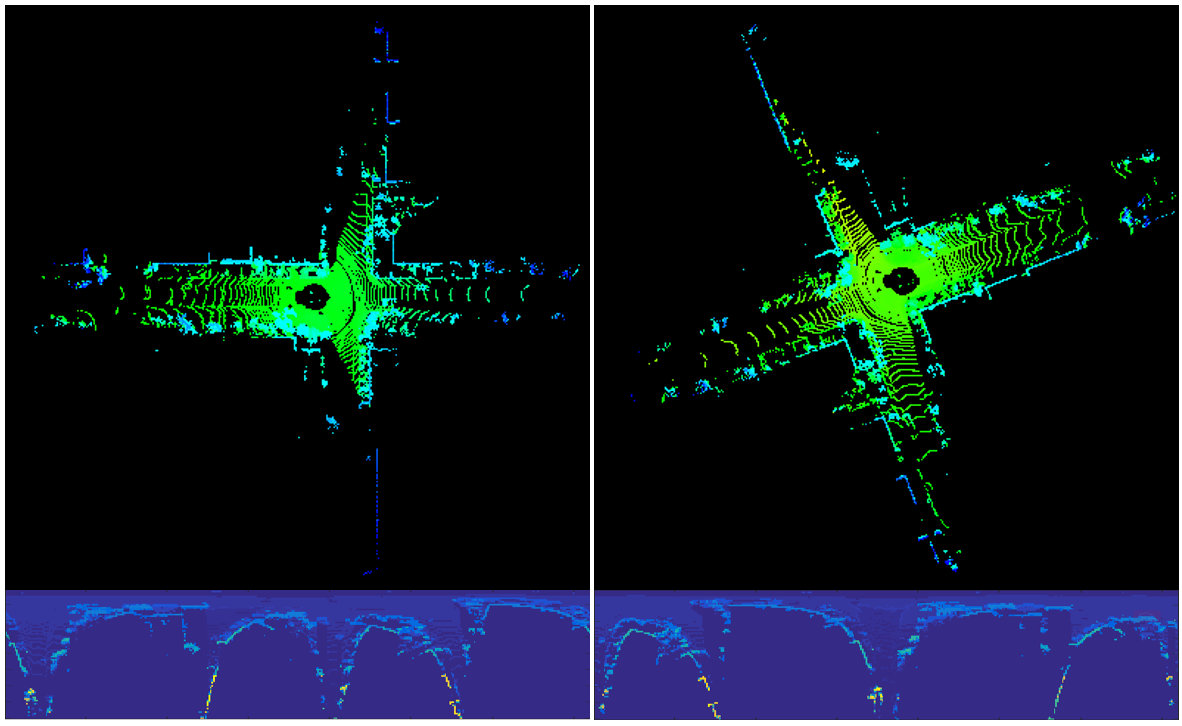}
		\caption{Two LiDAR-Iris images extracted from the bird's eye views of two LiDAR point clouds in a 3D (x,y,yaw) pose space, respectively. Apparantly, the two LiDAR-Iris images are subject to a cyclic translation when a loop-closure event occurs, with the LiDAR's pose being subject to a rotation. }
		\label{lidarIris}
	\end{figure}

	We present a global descriptor for a LiDAR point clould, LiDAR Iris, for fast and accurate loop closure detection. 
	The name of our LiDAR discriptor was originated from the society of person identification based on human's iris signature. As shown in Fig.\ref{iris}, the commonly adopted Daugman’s Rubber Sheet Model \cite{Daugman} is used to remap each point within the iris region to a pair of polar coordinates (\textbf{r},$\theta$), where \textbf{r} is on the interval [$r_1$,$r_2$], and $\theta$ represents the angle in the range [0,2$\pi$]. 
	We observe the similar characteristics between the bird's eye view of a LiDAR point cloud and an iris image of a human. Both can be represented in a polar-coordinate frame, and be transformed into a signature image. Fig.\ref{lidarIris} shows the bird's eye views of two LiDAR point clouds, and two extracted LiDAR-Iris images, respectively, based on the Daugman’s Rubber Sheet Model. 
	
	With the extracted LiDAR-Iris image representation, we generate a binary signature image for each point cloud by performing several LoG-Gabor filtering and thresholding operations on the LiDAR-Iris image. 
	For two binary signature images, the similarity of them can be calculated by the Hamming distance. 
	Our LiDAR-Iris method can deal with pose variation problem in LiDAR-based loop-closure detection.

	\section{RELATED WORK}
	
	Compared with visual loop-closure detection, LiDAR-based loop-closure detection has received continually increasing attention due to its robustness against illumination changes and high accuracy in localization. 
	
	In general, loop detection using 3D data can roughly be categorized into four main classes. The first category is point-to-point matching, which operates directly on point clouds. Popular
	methods in this category include the iterative closest points (ICP) \cite{ICP} and its variants \cite{ICPv,ICPv2}, in the case when two point clouds are already roughly aligned. 
	
	To improve matching capability and robustness, the second category applies corner (keypoint) detector to the 3D point cloud, extracts a local descriptor from each keypoint location, and conducts scene matching based on a bag-of-words (BoW) model. Many keypoint detection methods have been proposed in the literature, such as 3D Sift\cite{3D-SIFT}, 3D Harris \cite{3D-Harris}, 3D-SURF\cite{3D-SURF}, intrinsic shape signatures (ISSs) \cite{ISS}, as well as descriptors such as SHOT \cite{shot} and B-SHOT\cite{B-SHOT}, etc. 
	
	However, the detection of distinctive keypoints with high repeatability remains a challenge in 3D point cloud analysis. One way to deal with this problem is to extract global descriptors (represented in the form of histograms) from point cloud, e.g., the point feature histogram \cite{PFH}, ensemble of shape functions  (ESF)\cite{ESF}, fast point feature histogram (FPFH) \cite{FPFH} and the viewpoint feature histogram (VFH)\cite{VFH}. Recently, a novel global 3D descriptor for loop detection, named multiview 2D projection (M2DP) \cite{M2DP}, is proposed. This descriptor  suffers from the lack of rotation-invariance, where a Principal Component Analysis (PCA) operation applied to align point cloud is not a robust way to achieve invariance to rotation.

	However, either the global or local descriptor matching continues to suffer from, respectively, the lack of descriptive power or the struggle with invariance. More recently, convolutional neural networks (CNNs) models are exploited to learn both feature descriptors\cite{Learning1,Learning2,semantic,SegMap} as well as metric for matching point cloud \cite{PointNetVLAD,LPD-Net,LocNet}. However, a severe limitation of these deep-learning based methods is that they need a tremendous amount of training data. Moreover, they do not generalize well when trained and applied on data with varying topographies or acquired under different conditions. 
	
	The most similar work to our method is the Scan-Context (SC) \cite{sc} for loop-closure detection, which also exploits 
	the expanded bird-eye view of LiDAR's point cloud. Our method is different in three aspects: First, we encode height information of surroundings as the pixel intensity of the LiDAR-Iris image. Second, we extract a discriminative 
	binary feature map from the LiDAR-Iris image for loop-closure detection. Third, our loop-closure detection step is rotation-invariant with respect to LiDAR's pose. In contrast, in the Scan-Context method, only the maximum-height information is encoded in their expanded images and also lacks the feature extraction step. In addition, the Scan-Context method is not rotation-invariant, where a brute-force matching scheme is adopted. 
	
	\section{LiDAR Iris}
	
	This section consists of four different parts: discretization and encoding of bird's eye view image, generation and binarization of LiDAR-Iris image.
	
	\subsection{Generation of LiDAR-Iris Image Representation}
	
	\begin{figure}
		\centering 
		\includegraphics[width=\linewidth]{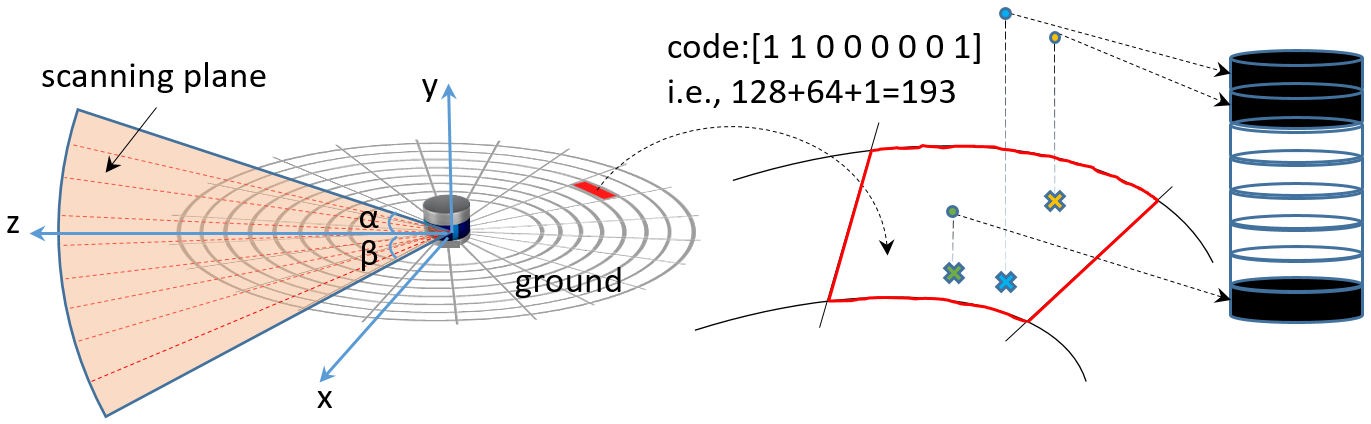}
		\caption{Illustration of encoding height information of surrounding objects into the LiDAR-Iris image.}
		\label{lidarIrisIntensity}
	\end{figure}

	Given a point cloud, we first project it to its bird's eye view, as shown in Fig.\ref{lidarIrisIntensity}. We keep a square of area of $k\times k$ $m^{2}$ as the valid sensing zone where the LiDAR's position is the center of the square. Typically, we set $k$ to 80$m$ in all of our experiments. The sensing square is discretized into 80 (radial direction)$\times$ 360 (angular direction) bins, $\emph{B}_i^{j}, i\in[1,80],j\in[1,360]$, with the angular resolution being 1$^{\circ}$ and radial resolution being 1$m$ (shown in Fig.\ref{lidarIrisIntensity}). 
	
	In order to fully represent the point cloud, one can adopt some feature extraction methods from points within each bin, such as height, range, reflection, ring and so on. For simplicity, we encode all the points falling within the same bin with an eight-bit binary code. If given an $N$-channel LiDAR sensor $\emph{L}$, its horizontal field-of-view (FOV) angle is 360$^{\circ}$
	and its vertical FOV is $\emph{V}$ degrees, with the vertical resolution being $V/N$ degrees.  As shown in Fig.\ref{lidarIrisIntensity}, if the largest and smallest pitch angle of all $N$ scan channels is $\alpha$ and $-\beta$, respectively, the highest and lowest point that the scan-line can reach is approximately $z\times \tan(\alpha)$ above ground and $-z\times \tan(\beta)$ under ground in theory ($z$ is the z-coordinate of the scanned point), respectively. In practice, the height of the lowest scanning point is set to the negative of the vehicle's height, i.e., ground plane level. We use $y_l$ and $y_h$ to represent the two quantities, respectively. 
	
	In this paper, we use the Velodyne HDL-64E (KITTI dataset) and VLP-16 (our own dataset) as the LiDAR sensors to validate our work. Typically, the $\alpha$ is $2^{\circ}$ for the HDL-64 and the $y_l$ and $y_h$ are -3$m$ and 5$m$, respectively, if $z$ is set to 80$m$ and the height of LiDAR is about 3$m$. Likewise, the $\alpha$ is $15^{\circ}$ for the VLP-16 and the $y_l$ and $y_h$ are -2$m$ and 22$m$, respectively, if $z$ is set to 80$m$ and the height of LiDAR is about 2$m$. 
	
	With these notions, we encode the set of points falling within each bin $\emph{B}_i^{j}$, denoted by $\emph{P}_i^{j}$, as follows. First, we linearly discretize the range between $y_l$ and $y_h$ into 8 bins, denoted by $\emph{y}_k$, $k\in[1,8]$. Each point of $\emph{P}_i^{j}$ is assigned into one of the bins based on the y-coordinate of each point. Afterwards, 
	$\emph{y}_k$, $k\in[1,8]$, is set to be zero if $\emph{y}_k$ is empty. Otherwise, $\emph{y}_k$ is set to be one. Thus, we are able to obtain an 8-bit binary code for each $\emph{B}_i^{j}$, as shown in Fig.\ref{lidarIrisIntensity}. The binary code within each bin $\emph{B}_i^{j}$ is turned into a decimal number between 0 and 255.
	
	Inspired by the work in iris recognition, we can expand the LiDAR's bird-eye view into an image strip, a.k.a. the LiDAR-Iris image, based on the Daugman’s Rubber Sheet Model \cite{Daugman}. The pixel's intensity of the LiDAR-Iris image is the decimal number calculated for each $\emph{B}_i^{j}$, and the size of the LiDAR-Iris image is i rows and j columns.   
	On the one hand, compared with the existing histogram-based global descriptors \cite{ESF} \cite{M2DP}, the proposed encoding procedure does not need to count the points in each bin, instead it proposes a more efficient bin encoding function for loop-closure detection. 
	On the other hand, our encoding procedure is fixed and does not require prior training like CNNs models. 
	Note that the obtained LiDAR-Iris images of the same geometrical location are up to a translation if we assume a 3D (x,y and yaw) pose space. In general, our method can be applied to detect loop-closure in 6D pose space if the LiDAR is combined with an IMU sensor. With the calibration of LiDAR and IMU, we can re-align point clouds so that the z-axis of LiDAR is identical with the gravity direction.

	\begin{figure}
		\centering 
		\includegraphics[width=\linewidth]{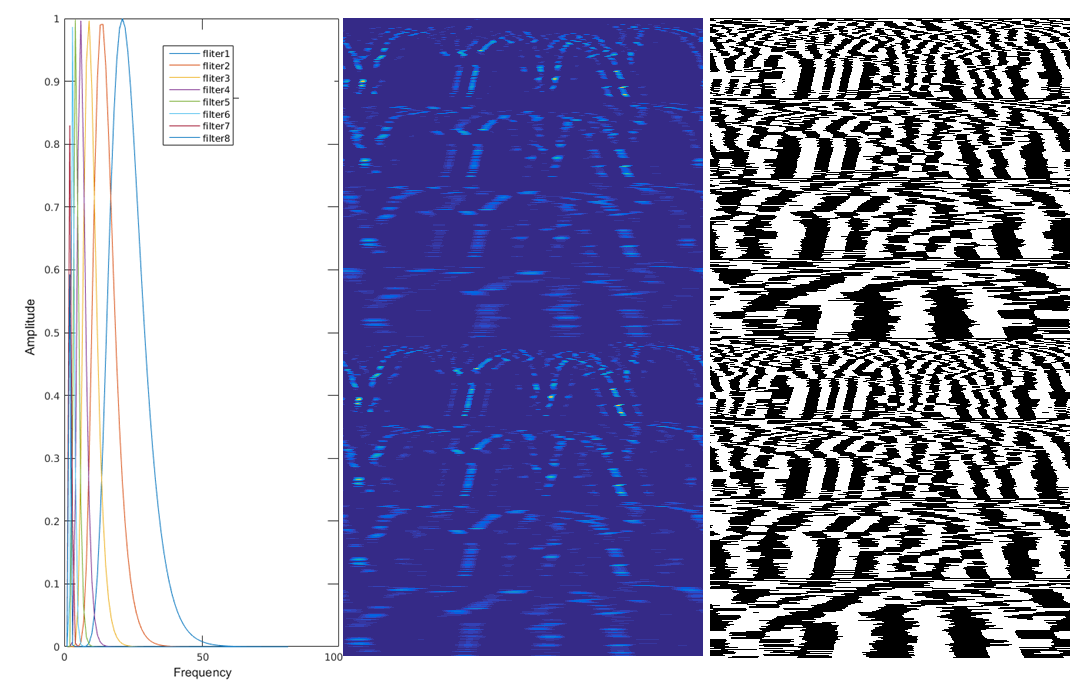}
		\caption{Left: eight LoG-Gabor filters. Middle: the real and imaginary parts of the convolutional response with four LoG-Gabor filters. Right: The binary feature map by thresholding on the convolutional response.}
		\label{gaborCode}
	\end{figure}

	In Fig.\ref{lidarIris}, the bottom two images are the LiDAR-Iris image representation of the two corresponding LiDAR point clouds. The two point clouds are collected when a robot was passing by the same geometrical location twice, and the collected point clouds are approximately subject to a rotation. Correspondingly, the two LiDAR-Iris images are mainly subject to a cyclic translation. 
	
	\subsection{Fourier transform for a translation-invariant LiDAR Iris}
	
\begin{figure}[t]
		\centering 
		\includegraphics[width=0.8\linewidth]{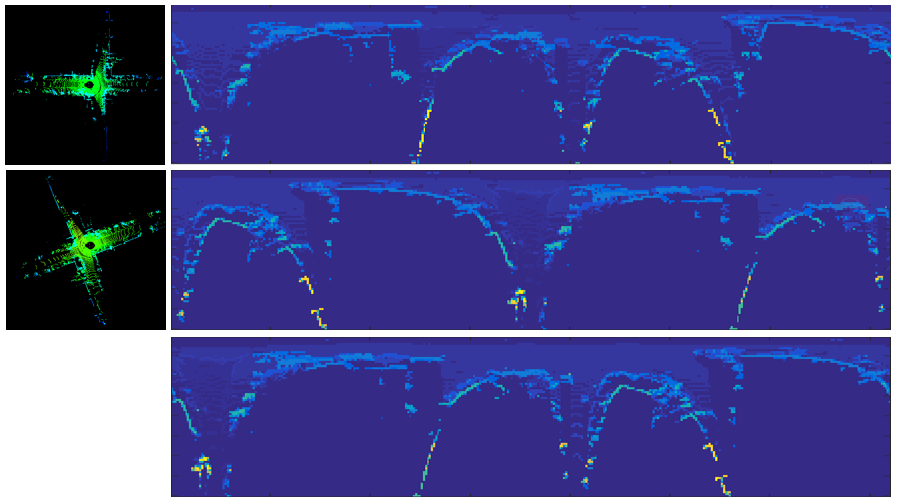}
		\caption{An example of achieving rotation invariance in matching two point clouds through alignment of the corresponding LiDAR-Iris images based on Fourier transform.}
		\label{Fourier}
	\end{figure}

	The translation variation shown above can cause a significant degradation in matching LiDAR-Iris images for LiDAR-based loop-closure detection. To deal with this problem, we adopt the Fourier transform to estimate the translation between two LiDAR-Iris images. The Fourier-based schemes are able to estimate large rotations, scalings, and translations. It should be noted that the rotation and scaling factor is irrelevant in our case. 
	The different orientations of the robot are expressed as the different rotations of the point cloud with the lidar's y-axis (see Fig.\ref{lidarIrisIntensity}). 
	The rotation of the point cloud corresponds to the horizontal translation on LiDAR-Iris image after the Fourier transform. 
	The robot's translation not only brings a vertical translation on the LiDAR-Iris image after the Fourier transform, but also causes a slight change in the intensity of the image pixels. 
	Our encoding method preserves the absolute internal structure of the point cloud with the bin as the smallest unit. 
	This improves the discrimination capability and is robust to the change in the intensity of the image pixels. 
	Therefore, we ignore the negligible changes in LiDAR-Iris images caused by the translation of the robot in a small range. 
	Fig.\ref{Fourier} gives an example of alignment based on Fourier transform, where the third row is a transformed version of the second LiDAR-Iris image. 
	
	Suppose that two LiDAR-Iris images $I_1$ and $I_2$ differ only by a shift ($\delta_x, \delta_y$) such that $I_1(x,y)$ = $I_2(x-\delta_x, y-\delta_y)$. The Fourier transform of $I_1$ and $I_2$ is related by 
	\begin{equation}
	\hat{I}_{1}(w_x,w_y)\dot e^{-i(w_{x}\delta_x+ w_{y}\delta_y)} = \hat{I}_{2}(w_x,w_y)
	\end{equation}
	Correspondingly, the normalized cross power spectrum is given by
	
	\begin{equation}
	\hat{Corr} = \frac{\hat{I}_{2}(w_x,w_y)}{\hat{I}_{1}(w_x,w_y)}==\frac{\hat{I}_{2}(w_x,w_y)\hat{I}_{1}(w_x,w_y)*}{|\hat{I}_{1}(w_x,w_y)\hat{I}_{1}(w_x,w_y)*|}=e^{-i(w_{x}\delta_x+ w_{y}\delta_y)}
	\end{equation}
	where $*$ indicates the complex conjugate. Taking the inverse Fourier transform $Corr(x,y) = F^{-1}(\hat{Corr})=\delta(x-\delta_x, y-\delta_y)$, meaning that $Corr(x,y)$ is nonzero only at 
	 ($\delta_x, \delta_y$) = $\arg\max_{x,y}\{Corr(x,y)\}$.

	\begin{figure}[t]
		\centering 
		\includegraphics[width=\linewidth]{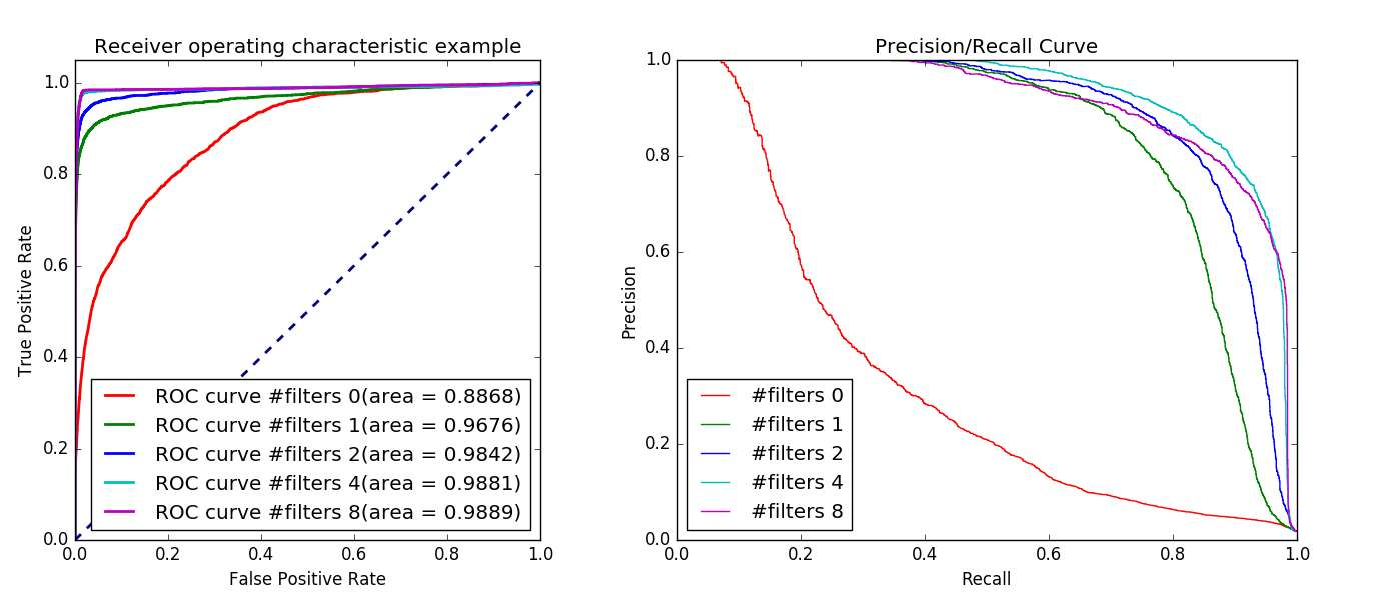}
		\caption{The loop-closure performance using different number of LoG-Gabor filters on a validation dataset. It shows that four LoG-Gabor filters can achieve best performance.}
		\label{gaborFilterScale}
	\end{figure}
	
	\subsection{Binary feature extraction with LoG-Gabor filters}
	To enhance the representation ability, we exploit LoG-Gabor filters to extract more features from LiDAR-Iris images.  LoG-Gabor filter can be used to decompose the data in the LiDAR-Iris region into components that
	appear at different resolutions, and it has the advantage over traditional Fourier
	transform in that the frequency data is localised, allowing features which occur at the same position and resolution to be matched up. 
	We only use 1D LoG-Gabor filters to ensure a real-time 
	capability of our method.  	
	A one dimensional Log-Gabor filter has the frequency response:
	\begin{equation}
		G(f) = \exp(\frac{-(\log(f/f_0))^2}{2(\log(\sigma/f_0))^2})
	\end{equation}
	where $f_0$ and $\sigma$ are the parameters of the filter. $f_0$ will give the center frequency of the filter. $\sigma$ affects the bandwidth of the filter.  It is useful to maintain the same shape while the frequency parameter is varied. To do this, the ratio $\sigma /f_{0}$ should remain constant.

	Eight 1D LoG-Gabor filters are exploited to convolve each row of the LiDAR-Iris image, where the wavelength of the filter is increased by the same factor, resulting in the real and the imaginary part for each filter. 
	As shown in Fig.\ref{gaborCode}, the first image shows the eight LoG-Gabor filters, and the second image shows the real and imaginary parts of the convolution response with the first four filters.

	Empirically, we have tried using a different number of LoG-Gabor filters for feature extraction, and have found that four LoG-Gabor filters can achieve best loop-closure detection accuracy at a low computational cost. Fig.\ref{gaborFilterScale} shows the accuracy that can be achieved on a validation dataset with different number of LoG-Gabor filters, where we obtain best results with the first four filters. Therefore, we only use the first four filters in obtaining our experimental results. 
	The convolutional responses with the four filters are turned into binary by a simple thresholding operation, and thus we stack them into a large binary feature map for each LiDAR-Iris image. For example, the third image of Fig.\ref{gaborCode} shows one binary feature map for a LiDAR-Iris image.

	\section{Loop-Closure Detection with LiDAR Iris}
	
	In a full SLAM method, loop-closure detection is an important step to trigger the backend optimization procedure to correct the already estimated pose and maps. To apply the LiDAR Iris to detect loops, we obtain a binary feature map with LiDAR-Iris representation for each image. Therefore, we can obtain a history database of LiDAR-Iris binary features for all keyframes that are saved when a robot is traversing in a scene. The distance between the LiDAR-Iris binary feature maps of the current keyframe and each of the history keyframes is calculated by Hamming distance. If the obtained Hamming distance is smaller than a threshold, it is regarded as a loop-closure event.

	\begin{figure*}
	   	\centering
	   	\includegraphics[width=0.8\linewidth]{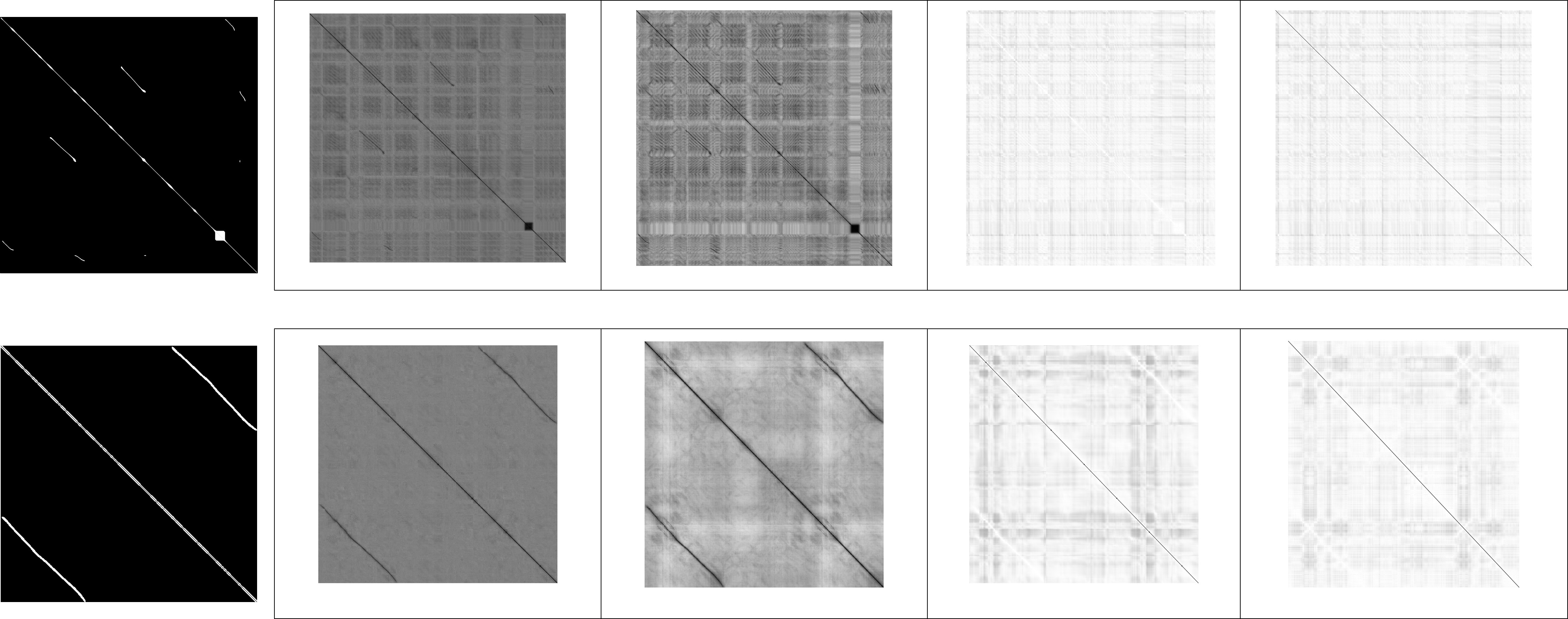}
	   	\caption{The affinity matrices obtained by the four compared methods on two sequences. The first row corresponds to the KITTI sequence 05, while the second row corresponding to our smaller scene. The far left is the ground truth affinity matrix. From left to right are the results of our approach, Scan-Context, M2DP and ESF, respectively. In our and Scan-Context's affinity matrices, the darker in the image, the higher the similarity of the two point clouds. In the affinity matrices of M2DP and ESF, the lighter the pixels, the more similar of the two point clouds.}
	   	\label{cm}
	\end{figure*}
    
	\section{Experiment}
	
	In this section, we compare the LiDAR-Iris method with a few other popular algorithms in loop-closure detection. 
	Since LiDAR Iris is a global descriptor, the performance of our method is compared to three other methods that extract global descriptors for a 3D point cloud. Specifcally, they are Scan-Context, M2DP and ESF. The codes of the three compared methods are available. The ESF method is implemented in the Point Cloud Library(PCL), and the Matlab codes of Scan-Context and M2DP can be downloaded from the author's website. All experiments are carried out on the same PC with an Intel i7-8550U CPU at 1.8GHZ and 8GB memory.

	\begin{figure}
		\centering 
		\includegraphics[width=0.8\linewidth]{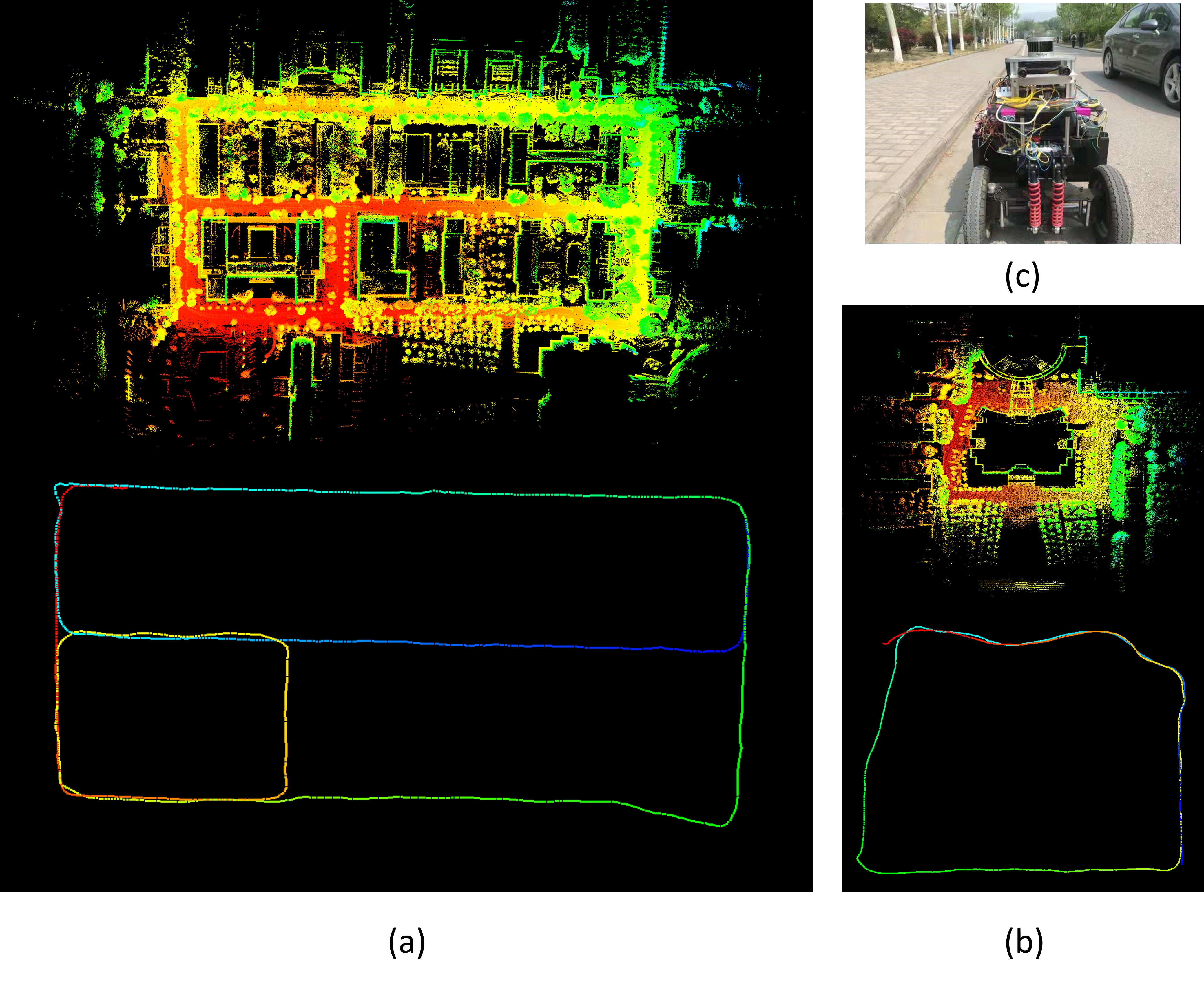}
		\caption{Our own VLP-16 dataset collected on our campus. In (a) and (b), the upper shows the reconstructed maps, and the lower shows the trajectories of the robot. }
		\label{map}
	\end{figure}

	\subsection{Dataset}
	
	All experimental results are obtained based on performance comparison on three KITTI odometry sequences and two collected on our campus. These datasets are considered diversity, such as the type of 3D LiDAR sensors (64 channels for Velodyne HDL-64E and 16 channels for Velodyne VLP-16) and the type of loops (eg., loop events occuring at places where robot moves in the same or opposite direction).
	
	\begin{figure}
		\centering
		\includegraphics[width=0.85\linewidth]{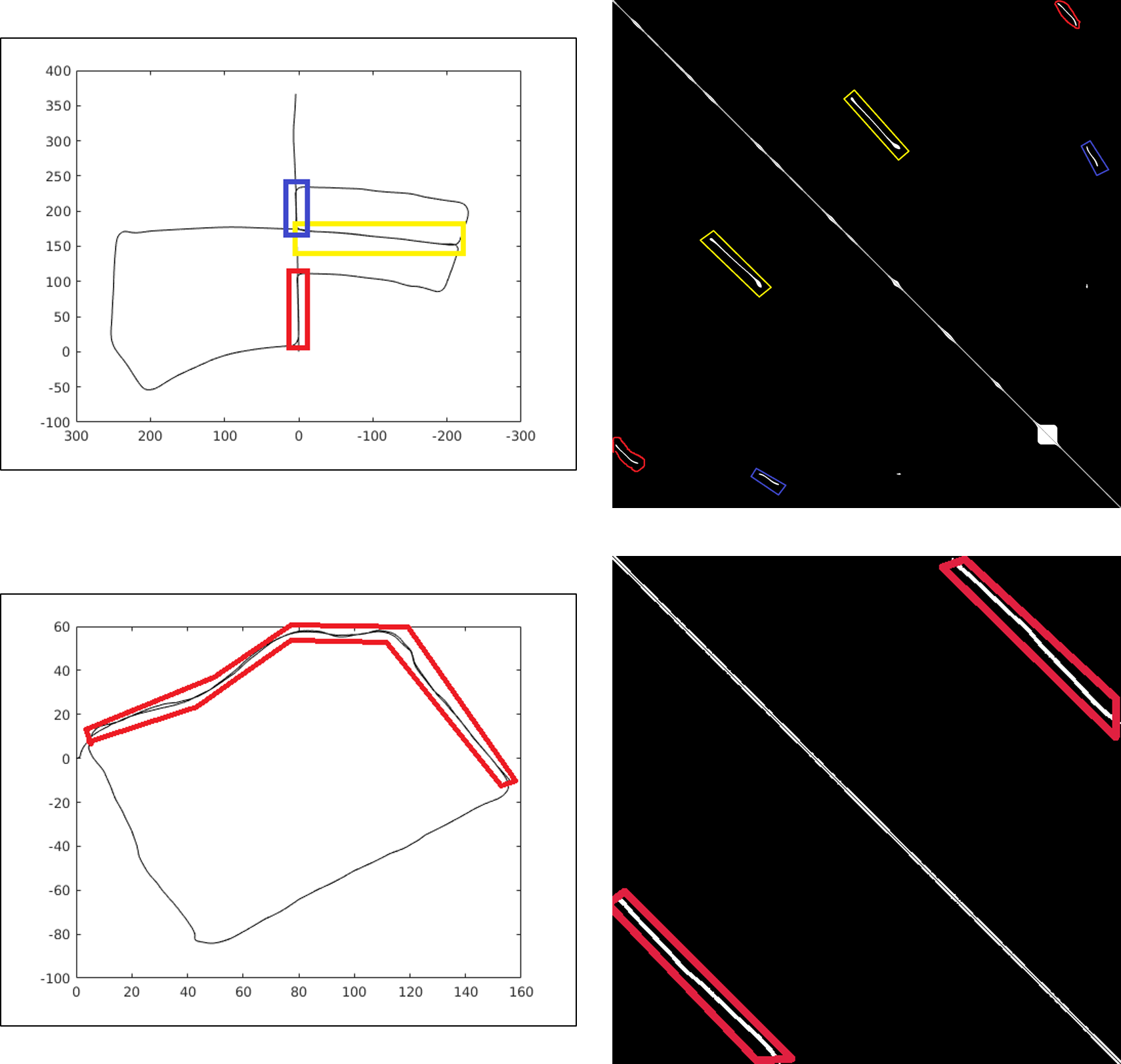}
		\caption{the loop-closure events areas occurred in the KITTI sequence 05 and our smaller scene. The left column is the trajectory, and the right column is the corresponding ground truth affinity matrices. We use the same colour to indicate the correspondence between trajectory and affinity matrix.}
		\label{tc}
	\end{figure}

	$KITTI$ $odometry$ $dataset$: Among the 11 sequences with the ground truth of pose (from 00 to 10), we selected three sequences 00, 05 and 08 which contain the largest number of loop-closure events. The sequence 08 has loop events only occuring at locations where robot/vehicle moves in the directions opposite to each other, and others have only loop events in the same direction. The scans of the KITTI dataset had been obtained from the Velodyne HDL-64E. Since the KITTI dataset provides scans with indexes, we use obtain loop-closure data easily.
	
	$Our$ $own$ $VLP-16$ $dataset$: We collected our own data on campus by using the Velodyne VLP-16 mounted on our mobile robot, Fig.\ref{map}(c). We selected two different-sized scenarios from our own VLP-16 dataset for the validation of our method. The smaller scene Fig.\ref{map}(b) has only loop events of the same direction, and the larger scene Fig.\ref{map}(a) has loop events of both the same and opposite directions. In order to get the ground truth of pose and location for our data, we use high-fidelity IMU/GPS to record the poses and locations of each LiDAR's frame. We only use the keyframes and their ground-truth locations in our experiment. Note that the distance between two keyframe locations is set to 1$m$.

	\begin{figure*}
		
		\centering
		\includegraphics[width=1\linewidth]{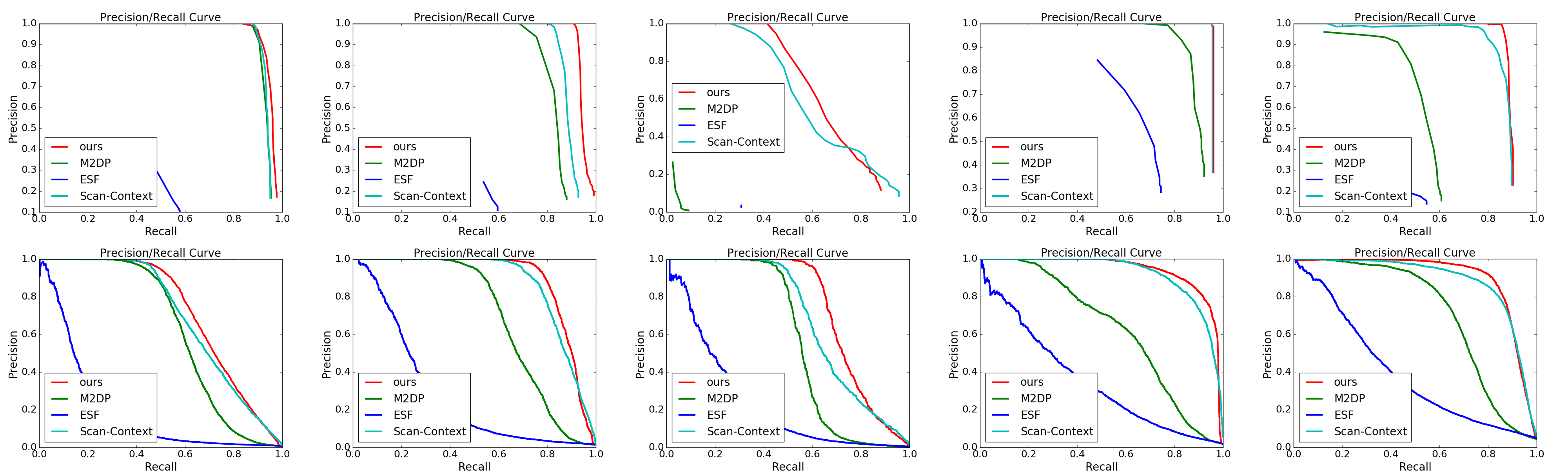}
		\caption{The precision-recall curves of different methods on different datasets. The first row and the second row represent the results under the protocols $protocolA$ and $protocolB$, respectively. From left to right are the results of KITTI 00, KITTI 05, KITTI 08, our smaller and larger scenes, respectively.}
		\label{pr}
	\end{figure*}

	\subsection{Experimental Settings}
	In order to  demonstrate the performance of our method thoroughly, we adopt two different protocols when evaluating the recall and precision of each compared method. 
	
	The first protocol, $protocol$ $A$, is a real one for loop-closure detection. Suppose that the keyframe for current location is $f_{kc}$, to find whether the current location has been traversed or not, we need to match $f_{kc}$ with all previous keyframes in the map database except the very close ones, e.g., the 30 keyframes ahead of the current one. By setting a threhold on the 
	feature distance, denoted by $d_f$, between $f_{kc}$ and the closest match in the database, we can predict whether $f_{kc}$ corresponds to an already traversed place or not. If the feature distance is no larger than $d_f$, $f_{kc}$ is predicted as a loop closure. Otherwise,  $f_{kc}$ is predicted as not a loop closure. 	
	To obtain the true-positive and recall rate, the prediction is further verified against the ground-truth. For example, if $f_{kc}$ is predicted as a loop-closure event, it is regarded as a true positive only if the ground truth distance between $f_{kc}$ and the closest match in the database is less than 4m. Note that the 4m-distance is set as default according to \cite{sc}.

	The second protocol, $protocol$ $B$, treats loop-closure detection as a place re-identification (re-ID) problem. First, we create the positive and negative pairs according to Euclidean distance between the two keyframes' locations. 
	For example, the $f_i$ and $f_j$ are two keyframes of the same sequence and their poses are $p_i$ and $p_j$, respectively. If ${||p_{i} - p_{j}||}_{2} \leq 4m$, the pair $(f_i, f_j)$ is a positive match pair, and otherwise a negative pair. By calculating pairwise feature distance for all $(f_i, f_j)$, we can get an affinity matrix for each sequence, shown in Fig.\ref{cm}. Likewise, by setting a threhold $d_f$ on the affinity matrix, we can obtain the true-positive and recall rate.

	\subsection{Performance Comparison}
	
	In all experiments in this paper, we set the parameters of Scan-Context as $N_s = 60$, $N_r = 20$ and $L_{max} = 80m$ used in their paper \cite{sc} and the default parameters of the available codes for M2DP and ESF. All methods use the raw point cloud without downsampling. 
	
	As shown in the top row of Fig.\ref{tc}, in the KITTI sequence 05, three segments of loop-closure events are labeled in different colors in the trajectory and the corresponding affinity matrix. The bottom row also highlights the loop-closure part in the shorter sequence of our own data. 
	Fig.\ref{cm} shows the affinity matrices obtained by the four compared methods on two exemplar sequences (KITTI 05 and the shorter one collected by us). 
	From left to right are the ground-truth results, LiDAR Iris, Scan-Context, M2DP and ESF, respectively. It is clear that the loop-closure areas can be detected effectively by the LiDAR-Iris and Scan-Context methods. Intuitively, Fig.\ref{cm} shows that the global descriptor generated by using our approach can more easily distinguish positive match pairs from negative match pairs. Although the affinity matrix obtained by Scan-Context can also find the loop-closure areas, several negative pairs also exhibit low matching values, which can more easily cause false positives. In contrast, M2DP and ESF are much worse than ours and the Scan-Context method.
	
	The performance of our method can also be validated from the precision-recall (PR) curve in Fig.\ref{pr}, where the top and bottom rows represent the results based on $protocol$ $A$ and $protocol$ $B$, respectively. 
	
	As shown in Fig.\ref{pr}, from left to right are the results of KITTI 00, KITTI 05, KITTI 08, our smaller scene and larger scene data, respectively. The ESF approach shows worst performance on all sequences under the two protocols. This algorithm strongly relies on the histogram and distinguish places only when the structure of the visible region is substantially different. 
	
	Under the $protocol$ $A$, M2DP reported high precision on the sequence that has only same-direction loop-closure events, which shows that M2DP can detect these loops correctly. However, when the sequence contains only opposite-direction loop-closure events (such as the KITTI 08) or has both same- and opposite-direction loop-closure events (our larger scene dataset), it fails to detect a loop-closure. The Scan-Context method can achieve promising results. In contrast, our approach demonstrates very competitive performance on the all five sequences, and achieves the best performance among the four compared methods. The superior performance of our method over the other compared ones originates in the merits of our method. First, the binary feature map of the LiDAR-Iris representation has very discriminative ability. Second, the translation-invariance achieved by the fourier transform can deal with the opposite loop-closure problem. Although the descriptor generated by Scan-Context can also achieve translation-invariance,  it uses a quite brute-force view alignment based matching. 
	
	Under the $protocol$ $B$, the number of negative pairs is much larger than that of positive pairs, and the number of total matching pairs to be predicted is much larger than that under the $protocol$ $A$. For example, in the KITTI sequence 00, there are 4541 bin files and 790 of them are true loops under the $protocol$ $A$. However, we will generate 68420 positive pairs and 20547720 negative pairs under the protocol $protocol$ $B$.
	Likewise, our method also achieves the best performance.

	\subsection{Computational Complexity}
	\begin{figure}
			\centering
			\includegraphics[width=0.93\linewidth]{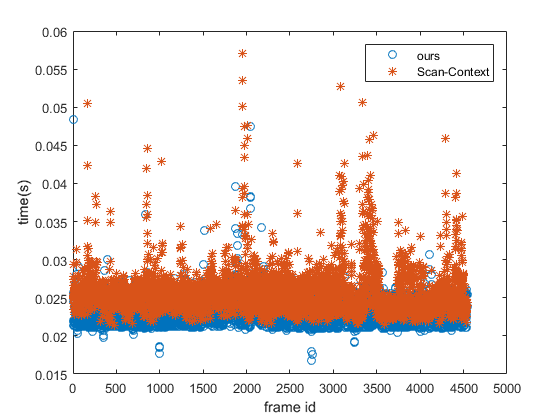}
			\caption{the computational complexity evaluated on KITTI sequence 00.}
			\label{time}
	\end{figure}

	We only compare our method with Scan-Context in terms of computational complexity on the KITTI sequence 00. Fig.\ref{time} shows that our method achieves better performance than the Scan-Context. The complexity of our method is evaluated in terms of the time spent on the binary feature extraction from LiDAR-Iris image and matching two binary feature maps, not including the time on generating LiDAR Iris. Similarly, the complexity of Scan-Context is evaluated in terms of the time spent on matching two Scan-Context images. Both methods are implemented in Matlab. 
	Specifically, we select a LiDAR frame from the KITTI sequence 00, and then calculate its distance from the same sequence. We obtain the average time it takes for each frame.  As shown in Fig.\ref{time}, the average computation time of our method is about 0.0231s and the time of Scan-Context is 0.0257s. It should be noted that when comparing the performance and computational complexity of our method and the Scan-Context, we did not set candidate parameters such as the Scan-Context 10 or 50 from the ring-key tree \cite{sc}, but compared all candidate key frames. Specifically, the PR curves in Fig.\ref{pr} show the highest performance that our method and the Scan-context can achieve. Fig.\ref{time} shows the time complexity of these two methods to match every two frames, so it is not affected by the number of candidates from the ring-key tree. 

	\section{conclusion}
	
	In this paper, we propose a global descriptor for a LiDAR point cloud, LiDAR Iris, summarizing a place as a binary signature image obtained after a couple of Gabor-filtering and thresholding operations on the LiDAR-Iris image representation. Compared to existing global descriptors using a point cloud, LiDAR Iris showed higher loop-closure detection performance across various datasets under two different protocol experiments.

	\bibliographystyle{IEEEtran}
    \bibliography{bibfile}

\end{document}